\documentclass[11pt,a4paper]{article}
\usepackage[hyperref]{acl2018}
\usepackage{times}
\usepackage{amsmath,amssymb}
\usepackage{booktabs}       
\usepackage{latexsym}
\usepackage{natbib}
\usepackage{graphicx}
\usepackage{graphics}
\usepackage{placeins}
\usepackage{float}
\usepackage{subfigure}
\usepackage{url}
\usepackage[backgroundcolor = White,textwidth=2cm]{todonotes}
\usepackage{xr}
\usepackage{etoolbox}
\makeatletter
\patchcmd\@combinedblfloats{\box\@outputbox}{\unvbox\@outputbox}{}{%
   \errmessage{\noexpand\@combinedblfloats could not be patched}%
}%
 \makeatother
\newcommand{\comment}[1]{}

\usepackage{hyperref}

\aclfinalcopy 
\def\aclpaperid{1001} 

\DeclareMathOperator{\E}{\mathbb{E}}
\title{Adversarial Contrastive Estimation}

\date{}

\begin{document}
\maketitle
\begin{abstract}
Learning by contrasting positive and negative samples is a general strategy adopted by many methods. Noise contrastive estimation (NCE) for word embeddings and translating embeddings for knowledge graphs are examples in NLP employing this approach. In this work, we view contrastive learning as an abstraction of all such methods and augment the negative sampler into a mixture distribution containing an adversarially learned sampler. The resulting adaptive sampler finds harder negative examples, which forces the main model to learn a better representation of the data. We evaluate our proposal on learning word embeddings, order embeddings and knowledge graph embeddings and observe both faster convergence and improved results on multiple metrics. 
\end{abstract}

\section{Introduction}
Many models learn by contrasting losses on observed positive examples with those on some fictitious negative examples, trying to decrease some score on positive ones while increasing it on negative ones. There are multiple reasons why such contrastive learning approach is needed. Computational tractability is one. For instance, instead of using softmax to predict a word for learning word embeddings, noise contrastive estimation (NCE) \cite{dyer2014notes,mnih2012fast} can be used in skip-gram or CBOW word embedding models \cite{gutmann2012noise, mikolov2013distributed,mnih2013learning,vaswani2013decoding}. Another reason is modeling need, as certain assumptions are best expressed as some score or energy in margin based or un-normalized probability models \cite{smith2005contrastive}. For example, modeling entity relations as translations or variants thereof in a vector space naturally leads to a distance-based score to be minimized for observed entity-relation-entity triplets \cite{transe}. 

Given a scoring function, the gradient of the model's parameters on observed positive examples can be readily computed, but the negative phase requires a design decision on how to sample data. In noise contrastive estimation for word embeddings, a negative example is formed by replacing a component of a positive pair by randomly selecting a sampled word from the vocabulary, resulting in a fictitious word-context pair which would be unlikely to actually exist in the dataset. This negative sampling by corruption approach is also used in learning knowledge graph embeddings \cite{transe,lin2015learning,transD,transH,complex,distmult,convE}, order embeddings \cite{OrderEmbedding15}, caption generation \cite{dai2017contrastive}, etc. 

Typically the corruption distribution is the same for all inputs like in skip-gram or CBOW NCE, rather than being a conditional distribution that takes into account information about the input sample under consideration. Furthermore, the corruption process usually only encodes a human prior as to what constitutes a hard negative sample, rather than being learned from data. For these two reasons, the simple fixed corruption process often yields only easy negative examples. Easy negatives are sub-optimal for learning discriminative representation as they do not force the model to find critical characteristics of observed positive data, which has been independently discovered in applications outside NLP previously \cite{shrivastava2016training}. Even if hard negatives are occasionally reached, the infrequency means slow convergence. 
Designing a more sophisticated corruption process could be fruitful, but requires costly trial-and-error by a human expert. 

In this work, we propose to augment the simple corruption noise process in various embedding models with an adversarially learned conditional distribution, forming a mixture negative sampler that adapts to the underlying data and the embedding model training progress. The resulting method is referred to as adversarial contrastive estimation (ACE). The adaptive conditional model engages in a minimax game with the primary embedding model, much like in Generative Adversarial Networks (GANs) \cite{NIPS2014_5423}, where a discriminator net (D), tries to distinguish samples produced by a generator (G) from real data \cite{goodfellow2014generative}. In ACE, the main model learns to distinguish between a real positive example and a negative sample selected by the mixture of a fixed NCE sampler and an adversarial generator. The main model and the generator takes alternating turns to update their parameters.
In fact, our method can be viewed as a conditional GAN \cite{ConditionalGan} on discrete inputs, with a mixture generator consisting of a learned and a fixed distribution, with additional techniques introduced to achieve stable and convergent training of embedding models. 

In our proposed ACE approach, the conditional sampler finds harder negatives than NCE, while being able to gracefully fall back to NCE whenever the generator cannot find hard negatives. We demonstrate the efficacy and generality of the proposed method on three different learning tasks, word embeddings \cite{mikolov2013distributed}, order embeddings \cite{OrderEmbedding15} and knowledge graph embeddings \cite{transD}.

\comment{
Contrastive learning approaches include max margin estimation used for SVM \cite{tsvetkov2016learning}, structural and latent structural svm \cite{sarawagi2008accurate}, \cite{yu2009learning}, as well as energy based models for supervised metric learning \cite{,hoffer2015deep}; contrastive divergence used for learning undirected graphical models such as RBM; Noise Contrastive Estimation (NCE) \cite{gutmann2010noise} and Negative Sampling \cite{goldberg2014word2vec} used as computationally efficient replacements for full softmax. From the perspective of energy based models, learning needs to push energy down on positive samples, while push up at any other points in space \cite{lecun2005loss}. In all of these cases, positive examples are observed and hence losses and gradient of losses can be evaluated easily on them, but negative examples are not. Different estimation methods make different choices about where to evaluate negative example losses. 
}


\section{Method}
\subsection{Background: contrastive learning}
\label{sec:contrastive}
In the most general form, 
our method applies to supervised learning problems with a contrastive objective of the following form:
\begin{equation}
\label{general_loss}
L(\omega) = \E_{p(x^+,y^+,y^-)} l_{\omega}(x^+,y^+,y^-)
\end{equation}
where $l_{\omega}(x^+,y^+,y^-)$ captures both the model with parameters $\omega$ and the loss that scores a positive tuple $(x^+,y^+)$ against a negative one $(x^+,y^-)$. $\E_{p(x^+,y^+,y^-)}(.)$ denotes expectation with respect to some joint distribution over positive and negative samples. 
Furthermore, by the law of total expectation, and the fact that given $x^+$, the negative sampling is not dependent on the positive label, i.e. $p(y^+,y^-|x^+) = p(y^+|x^+)p(y^-|x^+)$, Eq.\ \ref{general_loss} can be re-written as 
\begin{equation}
\label{general_loss_tot_E}
  \E_{p(x^+)}\lbrack\E_{p(y^+|x^+)p(y^-|x^+)} l_{\omega}(x^+,y^+,y^-)\rbrack
\end{equation}
\subsubsection*{Separable loss}
In the case where the loss decomposes into a sum of scores on positive and negative tuples such as $l_{\omega}(x^+,y^+,y^-) = s_{\omega}\left(x^+,y^+\right) - \tilde{s}_{\omega}\left(x^+,y^-\right)$, then Expression.\ \ref{general_loss_tot_E} becomes
\begin{equation}
\label{separable_loss}
\E_{p^+(x)}\lbrack \E_{p^{+}(y|x)}s_{\omega}\left(x,y\right)
- \E_{p^-(y|x)}\tilde{s}_{\omega}\left(x,y\right)\rbrack
\end{equation}
where we moved the $+$ and $-$ to $p$ for notational brevity. Learning by stochastic gradient descent aims to adjust $\omega$ to pushing down $s_{\omega}\left(x,y\right)$ on samples from $p^+$ while pushing up $\tilde{s}_{\omega}\left(x,y\right)$ on samples from $p^-$. Note that for generality, the scoring function for negative samples, denoted by $\tilde{s}_\omega$, could be slightly different from $s_\omega$. For instance, $\tilde{s}$ could contain a margin as in the case of Order Embeddings in Sec.\ \ref{sec:order_embd}. 

\subsubsection*{Non separable loss}
Eq.\ \ref{general_loss} is the general form that we would like to consider because for certain problems, the loss function cannot be separated into sums of terms containing only positive $(x^+,y^+)$ and terms with negatives $(x^+,y^-)$. 
An example of such a non-separable loss is the triplet ranking loss \cite{schroff2015facenet}: 
$l_{\omega} = \max(0, \eta + s_{\omega}\left(x^+,y^+\right) - s_{\omega}\left(x^+,y^-\right))$, which does not decompose due to the rectification.

\subsubsection*{Noise contrastive estimation}
The typical NCE approach in tasks such as word embeddings \cite{mikolov2013distributed}, order embeddings \cite{OrderEmbedding15}, and knowledge graph embeddings can be viewed as a special case of Eq.\ \ref{general_loss_tot_E} by taking $p(y^-|x^+)$ to be some unconditional $p_{nce}{(y)}$. 

This leads to efficient computation during training, however, $p_{nce}{(y)}$ sacrifices the sampling efficiency of learning as the negatives produced using a fixed distribution are not tailored toward $x^+$, and as a result are not necessarily hard negative examples. Thus, the model is not forced to discover discriminative representation of observed positive data. As training progresses, more and more negative examples are correctly learned, the probability of drawing a hard negative example diminishes further, causing slow convergence.

\subsection{Adversarial mixture noise}
To remedy the above mentioned problem of a fixed unconditional negative sampler, 
we propose to augment it into a mixture one, $\lambda p_{nce}{(y)} + (1-\lambda) g_{\theta}{(y|x)}$, where $g_\theta$ is a conditional distribution with a learnable parameter $\theta$ and $\lambda$ is a hyperparameter. The objective in Expression.\ \ref{general_loss_tot_E} can then be written as (conditioned on $x$ for notational brevity):
\begin{align}
\label{ace_value_general}
&L(\omega, \theta; x) = \lambda\E_{p(y^+|x)p_{nce}(y^-)} l_{\omega}(x,y^+,y^-) \nonumber \\
&+(1-\lambda)\E_{p(y^+|x)g_\theta(y^-|x)}l_{\omega}(x,y^+,y^-)
\end{align}
We learn $(\omega, \theta)$ in a GAN-style minimax game:
\begin{equation}
\min_\omega \max_\theta V(\omega, \theta) = \min_\omega \max_\theta \E_{p^{+}{(x)}} L(\omega, \theta; x)
\end{equation}
The embedding model behind $l_{\omega}(x,y^+,y^-)$ is similar to the discriminator in (conditional) GAN (or critic in Wasserstein \citep{arjovsky2017wasserstein} or Energy-based GAN \citep{zhao2016energy}, while $g_\theta(y|x)$ acts as the generator. Henceforth, we will use the term discriminator (D) and embedding model interchangeably, and refer to $g_\theta$ as the generator.

\subsection{Learning the generator}

There is one important distinction to typical GAN: $g_\theta(y|x)$ defines a categorical distribution over possible $y$ values, and samples are drawn accordingly; in contrast to typical GAN over continuous data space such as images, where samples are generated by an implicit generative model that warps noise vectors into data points. Due to the discrete sampling step, $g_\theta$ cannot learn by receiving gradient through the discriminator. One possible solution is to use the Gumbel-softmax reparametrization trick \cite{jang2016categorical, maddison2016concrete}, which gives a differentiable approximation. However, this differentiability comes at the cost of drawing $N$ Gumbel samples per each categorical sample, where $N$ is the number of categories. For word embeddings, $N$ is the vocabulary size, and for knowledge graph embeddings, $N$ is the number of entities, both leading to infeasible computational requirements. 

Instead, we use the REINFORCE \citep{williams1992simple} gradient estimator for $\nabla_{\theta}L(\theta,x)$: 
\begin{equation}
\label{reinfornce_general}
(1\!-\!\lambda)\E\left[-l_{\omega}(x,y^+,y^-)\nabla_{\theta}\log(g_\theta(y^-|x))\right]
\end{equation}
where the expectation $\E$ is with respect to $p(y^+,y^-|x)={p(y^+|x)g_\theta(y^-|x)}$, and 
the discriminator loss $l_{\omega}(x,y^+,y^-)$ acts as the reward. 

With a separable loss, the (conditional) value function of the minimax game is:
\begin{align}
\label{L_adv_expanded}
& L(\omega, \theta; x) = \E_{p^{+}(y|x)}s_{\omega}\left(x,y\right) \nonumber \\ 
 &- \E_{p_{nce}(y)}\tilde{s}_{\omega}\left(x,y\right) - \E_{g_\theta(y|x)}\tilde{s}_{\omega}\left(x,y\right)
\end{align}
and only the last term depends on the generator parameter $\omega$. 
Hence, with a separable loss, the reward is $-\tilde{s}(x^+,y^-)$.
This reduction does not happen with a non-separable loss, and we have to use $l_{\omega}(x,y^+,y^-)$. 

\subsection{Entropy and training stability}
\label{sec:entropy}
GAN training can suffer from instability and degeneracy where the generator probability mass collapses to a few modes or points. Much work has been done to stabilize GAN training in the continuous case \cite{arjovsky2017wasserstein,gulrajani2017improved, Cao2018Improving}. In ACE, if the generator $g_\theta$ probability mass collapses to a few candidates, then after the discriminator successfully learns about these negatives, $g_\theta$ cannot adapt to select new hard negatives, because the REINFORCE gradient estimator Eq.\ \ref{reinfornce_general} relies on $g_\theta$ being able to explore other candidates during sampling. 
Therefore, if the $g_\theta$ probability mass collapses, instead of leading to oscillation as in typical GAN, the min-max game in ACE reaches an equilibrium where the discriminator wins and $g_\theta$ can no longer adapt, then ACE falls back to NCE since the negative sampler has another mixture component from NCE. 

This behavior of gracefully falling back to NCE is more desirable than the alternative of stalled training if $p^-(y|x)$ does not have a simple $p_{nce}$ mixture component. However, we would still like to avoid such collapse, as the adversarial samples provide greater learning signals than NCE samples. To this end, we propose to use a regularizer to encourage the categorical distribution $g_\theta(y|x)$ to have high entropy. In order to make the the regularizer interpretable and its hyperparameters easy to tune, we design the following form:
\begin{equation}
R_{ent}(x) = \textnormal{max}(0, c-H(g_\theta(y|x)))
\end{equation}
where $H(g_\theta(y|x))$ is the entropy of the categorical distribution $g_\theta(y|x)$, and $c=\log(k)$ is the entropy of a uniform distribution over $k$ choices, and $k$ is a hyper-parameter. Intuitively, $R_{ent}$ expresses the prior that the generator should spread its mass over more than $k$ choices for each $x$. 

\subsection{Handling false negatives}
\label{sec:false_neg}
During negative sampling, $p^-(y|x)$ could actually produce $y$ that forms a positive pair that exists in the training set, i.e., a false negative. This possibility exists in NCE already, but since $p_{nce}$ is not adaptive, the probability of sampling a false negative is low. Hence in NCE, the score on this false negative (true observation) pair is pushed up less in the negative term than in the positive term. 

However, with the adaptive sampler, $g_\omega(y|x)$, false negatives become a much more severe issue. $g_\omega(y|x)$ can learn to concentrate its mass on a few false negatives, significantly canceling the learning of those observations in the positive phase. The entropy regularization reduces this problem as it forces the generator to spread its mass, hence reducing the chance of a false negative. 

To further alleviate this problem, whenever computationally feasible, we apply an additional two-step technique. First, we maintain a hash map of the training data in memory, and use it to efficiently detect if a negative sample $(x^+,y^-)$ is an actual observation. If so, its contribution to the loss is given a zero weight in $\omega$ learning step. Second, to update $\theta$ in the generator learning step, the reward for false negative samples are replaced by a large penalty, so that the REINFORCE gradient update would steer $g_\theta$ away from those samples. The second step is needed to prevent null computation where $g_\theta$ learns to sample false negatives which are subsequently ignored by the discriminator update for $\omega$.

\subsection{Variance Reduction}
The basic REINFORCE gradient estimator is poised with high variance, so in practice one often needs to apply variance reduction techniques. The most basic form of variance reduction is to subtract a baseline from the reward. As long as the baseline is not a function of actions (i.e., samples $y^-$ being drawn), the REINFORCE gradient estimator remains unbiased. More advanced gradient estimators exist that also reduce variance \cite{grathwohl2017backpropagation,NIPS2017_6856,liu2018actiondependent}, but for simplicity we use the self-critical baseline method \cite{rennie2016self}, where the baseline is $b(x) = l_\omega(y^+,y^\star,x)$, or $b(x) = -\tilde{s}_\omega(y^\star,x)$ in the separable loss case, and $y^\star = \text{argmax}_i g_\theta(y_i|x) $.
In other words, the baseline is the reward of the most likely sample according to the generator.

\subsection{Improving exploration in $\boldmath{g_\theta}$ by leveraging NCE samples}
In Sec.\ \ref{sec:entropy} we touched on the need for sufficient exploration in $g_\theta$. It is possible to also leverage negative samples from NCE to help the generator learn. This is essentially off-policy exploration in reinforcement learning since NCE samples are not drawn according to $g_\theta(y|x)$. The generator learning can use importance re-weighting to leverage those samples. The resulting REINFORCE gradient estimator is basically the same as Eq.\ \ref{reinfornce_general} except that the rewards are reweighted by $g_\theta(y^-|x)/p_{nce}(y^-)$, and the expectation is with respect to $p(y^+|x)p_{nce}(y^-)$.
This additional off-policy learning term provides gradient information for generator learning if $g_\theta(y^-|x)$ is not zero, meaning that for it to be effective in helping exploration, the generator cannot be collapsed at the first place. Hence, in practice, this term is only used to further help on top of the entropy regularization, but it does not replace it.


\section{Related Work}
\citet{smith2005contrastive} proposed contrastive estimation as a way for unsupervised learning of log-linear models by taking implicit evidence from user-defined neighborhoods around observed datapoints.
\citet{gutmann2010noise} introduced NCE as an alternative to the hierarchical softmax. In the works of \citet{mnih2012fast} and \citet{mnih2013learning}, NCE is applied to log-bilinear models and \citet{vaswani2013decoding} applied NCE to neural probabilistic language models \citep{Yoshua2003}. Compared to these previous NCE methods that rely on simple fixed sampling heuristics, ACE uses an adaptive sampler that produces harder negatives.

In the domain of max-margin estimation for structured prediction \citep{taskar2005learning}, loss augmented MAP inference plays the role of finding hard negatives (the hardest). However, this inference is only tractable in a limited class of models such structured SVM \citep{tsochantaridis2005large}. Compared to those models that use exact maximization to find the hardest negative configuration each time, the generator in ACE can be viewed as learning an approximate amortized inference network. Concurrently to this work, \citet{tu2018learning} proposes a very similar framework, using a learned inference network for Structured prediction energy networks (SPEN) \cite{belanger2016structured}.

Concurrent with our work, there have been other interests in applying the GAN to NLP problems \citep{fedus2018maskgan,wang2018incorporating,cai2017kbgan}. Knowledge graph models naturally lend to a GAN setup, and has been the subject of study in \citet{wang2018incorporating} and \citet{cai2017kbgan}. These two concurrent works are most closely related to one of the three tasks on which we study ACE in this work. Besides a more general formulation that applies to problems beyond those considered in \citet{wang2018incorporating} and \citet{cai2017kbgan}, the techniques introduced in our work on handling false negatives and entropy regularization lead to improved experimental results as shown in Sec.\ \ref{sec:wn}.

\section{Application of ACE on three tasks}
\subsection{Word Embeddings}
Word embeddings learn a vector representation of words from co-occurrences in a text corpus. NCE casts this learning problem as a binary classification where the model tries to distinguish positive word and context pairs, from negative noise samples composed of word and false context pairs. The NCE objective in Skip-gram \citep{mikolov2013distributed} for word embeddings is a separable loss of the form:
\vspace*{-.2cm}
\begin{equation}
\begin{split}
L = -\sum_{w_t \in V}[\log p(y=1|w_t,w^+_c)
\\ + \sum^K_{c=1} \log p(y=0|w_t,w^-_c)]
\end{split}
\end{equation}

Here, $w^+_c$ is sampled from the set of true contexts and $w^-_c \sim Q$ is sampled $k$ times from a fixed noise distribution. \citet{mikolov2013distributed} introduced a further simplification of NCE, called ``Negative Sampling" \cite{dyer2014notes}. With respect to our ACE framework, the difference between NCE and Negative Sampling is inconsequential, so we continue the discussion using NCE. A drawback of this sampling scheme is that it favors more common words as context. Another issue is that the negative context words are sampled in the same way, rather than tailored toward the actual target word. To apply ACE to this problem we first define the value function for the minimax game, $V(D,G)$, as follows:
\begin{equation}
\begin{split}
V(D,G) = \E_{p^+(w_c)}[\log D(w_c,w_t)] \\
- \E_{p_{nce}(w_c)}[-\log (1- D(w_c,w_t))] \\
- \E_{g_\theta(w_c|w_t)}[-\log(1- D(w_c,w_t))]
\end{split}
\end{equation}
with $D=p(y=1|w_t,w_c)$ and $G = g_\theta(w_c|w_t)$.
\subsubsection*{Implementation details}
For our experiments, we train all our models on a single pass of the May 2017 dump of the English Wikipedia with lowercased unigrams. The vocabulary size is restricted to the top $150k$ most frequent words when training from scratch while for finetuning we use the same vocabulary as \citet{pennington2014glove}, which is $400k$ of the most frequent words. We use $5$ NCE samples for each positive sample and $1$ adversarial sample in a window size of $10$ and the same positive subsampling scheme proposed by \citet{mikolov2013distributed}. Learning for both G and D uses Adam \cite{kingma2014adam} optimizer with its default parameters. Our conditional discriminator is modeled using the Skip-Gram architecture, which is a two layer neural network with a linear mapping between the layers. The generator network consists of an embedding layer followed by two small hidden layers, followed by an output softmax layer. The first layer of the generator shares its weights with the second embedding layer in the discriminator network, which we find really speeds up convergence as the generator does not have to relearn its own set of embeddings. The difference between the discriminator and generator is that a sigmoid nonlinearity is used after the second layer in the discriminator, while in the generator, a softmax layer is used to define a categorical distribution over negative word candidates. We find that controlling the generator entropy is critical for finetuning experiments as otherwise the generator collapses to its favorite negative sample. The word embeddings are taken to be the first dense matrix in the discriminator.

\subsection{Order Embeddings Hypernym Prediction}
\label{sec:order_embd}
As introduced in \citet{OrderEmbedding15}, ordered representations over hierarchy can be learned by order embeddings. An example task for such ordered representation is hypernym prediction. A hypernym pair is a pair of concepts where the first concept is a specialization or an instance of the second. 

For completeness, we briefly describe order embeddings, then analyze ACE on the hypernym prediction task.
In order embeddings, each entity is represented by a vector in $\mathbb{R}^N$, the score for a positive ordered pair of entities $(x, y)$ is defined by $s_{\omega}(x, y) = || max(0, y-x)||^2 $ and, score for a negative ordered pair $(x^+, y^-)$ is defined by  $\tilde{s}_{\omega}(x^+, y^-) = \max\{0, \eta - s(x^+, y^-)\} $, where is $\eta$ is the margin. Let $f(u)$ be the embedding function which takes an entity as input and outputs an embedding vector. We define $P$ as a set of positive pairs and $N$ as negative pairs, the separable loss function for order embedding task is defined by:
\begin{equation}
L \!=\! \!\sum_{(u,v) \in P}s_{\omega}(f(u), f(v))) + \! \sum_{(u,v) \in N}\tilde{s}(f(u), f(v))
\end{equation}

\subsubsection*{Implementation details}

Our generator for this task is just a linear fully connected softmax layer, taking an embedding vector from discriminator as input and outputting a categorical distribution over the entity set. For the discriminator, we inherit all model setting from \citet{OrderEmbedding15}: we use 50 dimensions hidden state and bash size 1000, a learning rate of 0.01 and the Adam optimizer. For the generator, we use a batch size of 1000, a learning rate 0.01 and the Adam optimizer. We apply weight decay with rate 0.1 and entropy loss regularization as described in Sec. \ref{sec:entropy}. We handle false negative as described in Sec. \ref{sec:false_neg}. After cross validation, variance reduction and leveraging NCE samples does not greatly affect the order embedding task.

\begin{figure*}[t]
\vspace{-3mm}
\hfill
\minipage{0.45\textwidth}
\minipage{0.5\textwidth}
\includegraphics[width=1.0\textwidth]{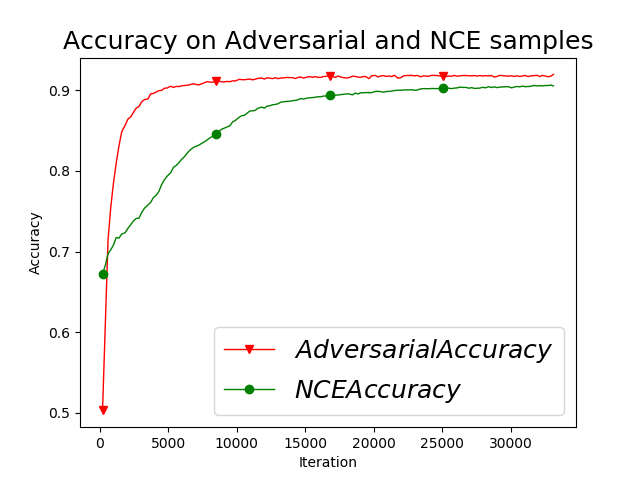}
\endminipage\hfill
\minipage{0.5\textwidth}
\includegraphics[width=1.0\textwidth]{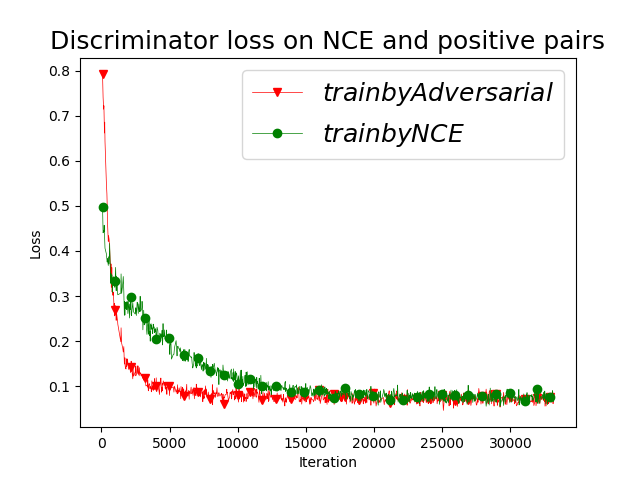}
\endminipage\hfill
\vspace{-1mm}
\caption{\small \textbf{Left}: Order embedding Accuracy plot. \textbf{Right}: Order embedding discriminator Loss plot on NCE sampled negative pairs and positive pairs.
 }
\endminipage
\hspace{2mm}
\minipage{0.45\textwidth}
\vspace{3mm}
\label{fig:order_plot}
\minipage{0.5\textwidth}
\vspace{-4mm}
\includegraphics[width=1.0\textwidth]{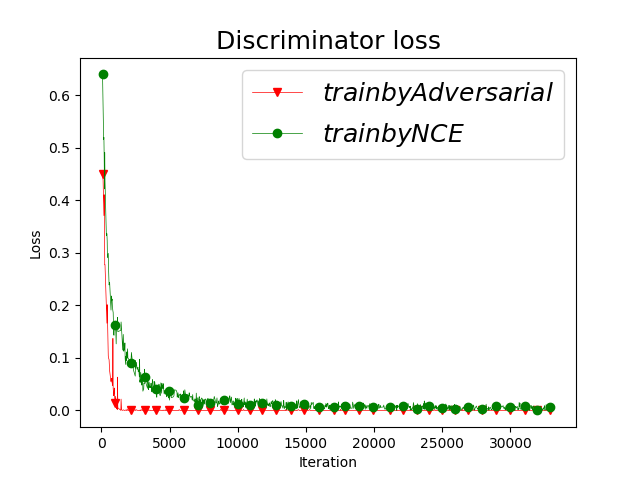}
\endminipage\hfill
\minipage{0.5\textwidth}
\vspace{-4mm}
\includegraphics[width=1.0\textwidth]{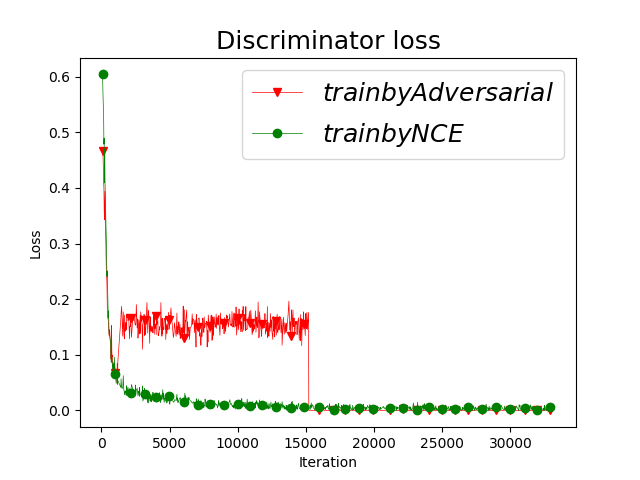}
\endminipage\hfill
\vspace{-1mm}
\caption{\small loss curve on NCE negative pairs and ACE negative pairs. \textbf{Left}: without entropy and weight decay. \textbf{Right}: with entropy and weight decay}
\label{fig:negative_pairs_loss_order_plot}
\endminipage\hfill
\vspace{-3mm}
\end{figure*}

\begin{figure*}
\minipage{0.65\textwidth}
\hfill
\minipage{0.5\textwidth}
\includegraphics[width=1.0\textwidth]{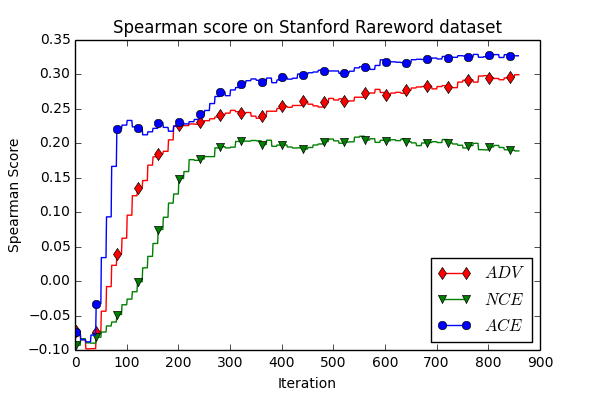}
\endminipage\hfill
\minipage{0.5\textwidth}
\includegraphics[width=1.0\textwidth]{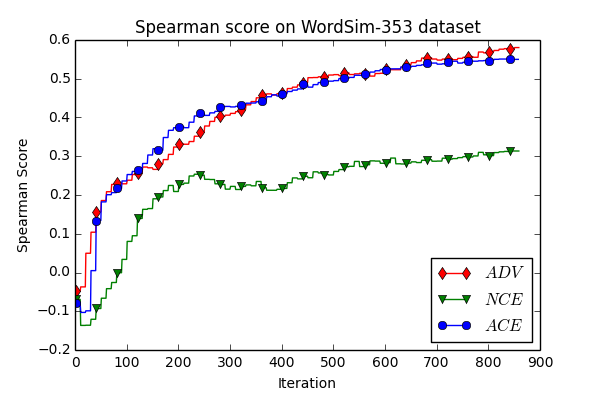}
\endminipage\hfill
\caption{\small \textbf{Left}: Rare Word, \textbf{Right}: WS353 similarity scores during the first epoch of training.}
\endminipage\hfill
\minipage{0.33\textwidth}
\hspace{-7mm}
\label{fig:word_embedding_plot2}
\vspace{3mm}\hfill
\includegraphics[width=1\textwidth]{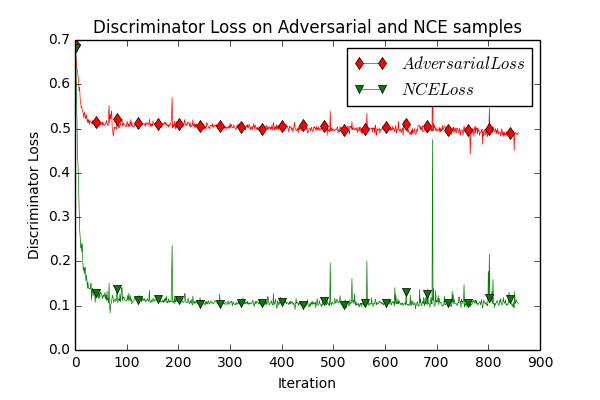}
\vspace{-3mm}
\caption{\small Training from scratch losses on the Discriminator}
\label{fig:word_embedding_plot1}
\endminipage\hfill
\vspace{-5mm}
\end{figure*}

\subsection{Knowledge Graph Embeddings}
\label{sec:kg}
Knowledge graphs contain entity and relation data of the form {\it{(head entity, relation, tail entity)}}, and the goal is to learn from observed positive entity relations and predict missing links (a.k.a. link prediction). There have been many works on knowledge graph embeddings, e.g. TransE \cite{transe}, TransR \cite{lin2015learning}, TransH \cite{transH}, TransD \cite{transD}, Complex \cite{complex}, DistMult \cite{distmult} and ConvE \cite{convE}. Many of them use a contrastive learning objective. Here we take TransD as an example, and modify its noise contrastive learning to ACE, and demonstrate significant improvement in sample efficiency and link prediction results. 

\subsubsection*{Implementation details}

Let a positive entity-relation-entity triplet be denoted by $\xi^+ = (h^+, r^+, t^+)$, and a negative triplet could either have its head or tail be a negative sample, i.e. $\xi^- = (h^-, r^+, t^+)$ or  $\xi^- = (h^+, r^+, t^-)$. In either case, the general formulation in Sec.\ \ref{sec:contrastive} still applies. The  non-separable loss function takes on the form:
\begin{equation}
l = \max(0, \eta + s_\omega(\xi^+) - s_\omega(\xi^-))
\end{equation}
The scoring rule is:
\begin{equation}
s = \lVert \mathbf{h}_\perp + \mathbf{r} - \mathbf{t}_\perp \rVert
\end{equation}
where $\mathbf{r}$ is the embedding vector for $r$, and $\mathbf{h}_\perp$ is projection of the embedding of $h$ onto the space of $\mathbf{r}$ by $\mathbf{h}_\perp = \mathbf{h} + \mathbf{r}_p\mathbf{h}^\top_p\mathbf{h}$, where $\mathbf{r}_p$ and $\mathbf{h}_p$ are projection parameters of the model. $\mathbf{t}_\perp$ is defined in a similar way through parameters $\mathbf{t}$, $\mathbf{t}_p$ and $\mathbf{r}_p$. 

The form of the generator $g_\theta(t^-|r^+,h^+)$ is chosen to be $f_\theta(\mathbf{h}_\perp,\mathbf{h}_\perp + \mathbf{r})$, where $f_\theta$ is a feedforward neural net that concatenates its two input arguments, then propagates through two hidden layers, followed by a final softmax output layer. As a function of $(r^+,h^+)$, $g_\theta$ shares parameter with the discriminator, as the inputs to $f_\theta$ are the embedding vectors. During generator learning, only $\theta$ is updated and the TransD model embedding parameters are frozen.


\section{Experiments}
We evaluate ACE with experiments on word embeddings, order embeddings, and knowledge graph embeddings tasks. In short, whenever the original learning objective is contrastive (all tasks except Glove fine-tuning) our results consistently show that ACE improves over NCE. In some cases, we include additional comparisons to the state-of-art results on the task to put the significance of such improvements in context: the generic ACE can often make a reasonable baseline competitive with SOTA methods that are optimized for the task. 

For word embeddings, we evaluate models trained from scratch as well as fine-tuned Glove models \cite{pennington2014glove} on word similarity tasks that consist of computing the similarity between word pairs where the ground truth is an average of human scores. We choose the Rare word dataset \cite{luong2013better} and WordSim-353 \cite{finkelstein2001placing} by virtue of our hypothesis that ACE learns better representations for both rare and frequent words. We also qualitatively evaluate ACE word embeddings by inspecting the nearest neighbors of selected words. 

For the hypernym prediction task, following \citet{OrderEmbedding15}, hypernym pairs are created from the WordNet hierarchy's transitive closure. We use the released random development split and test split from \citet{OrderEmbedding15}, which both contain 4000 edges.

For knowledge graph embeddings, we use TransD \cite{transD} as our base model, and perform ablation study to analyze the behavior of ACE with various add-on features, and confirm that entropy regularization is crucial for good performance in ACE. We also obtain link prediction results that are competitive or superior to the state-of-arts on the WN18 dataset \cite{bordes2014semantic}.

\begin{table*}[ht]
	\small
	\centering
	\begin{tabular}[t]{llllll}
    \toprule
	&Queen&King&Computer&Man&Woman\\
\hline
	Skip-Gram NCE Top 5
    &princess &prince &computers &woman &girl \\
    &king &queen &computing &boy &man \\
    &empress &kings &software &girl &prostitute \\
    &pxqueen &emperor &microcomputer &stranger &person \\
    &monarch &monarch &mainframe &person &divorcee \\
     \midrule
    Skip-Gram NCE Top 45-50
    &sambiria &eraric &hypercard &angiomata &suitor \\
    &phongsri &mumbere &neurotechnology &someone &nymphomaniac \\
    &safrit &empress &lgp &bespectacled &barmaid \\
    &mcelvoy &saxonvm &pcs &hero &redheaded \\
    &tsarina &pretender &keystroke &clown &jew  \\

    \midrule
	Skip-Gram ACE Top 5
    &princess &prince &software &woman &girl \\
    &prince & vi &computers &girl &herself \\
    &elizabeth &kings &applications &tells &man  \\
    &duke &duke &computing &dead &lover \\
    &consort &iii &hardware &boy &tells \\
    \midrule
    Skip-Gram ACE Top 45-50
    &baron  &earl &files &kid &aunt \\
    &abbey &holy &information &told &maid \\
    &throne&cardinal &device &revenge &wife \\
    &marie &aragon &design &magic &lady \\
    &victoria &princes & compatible & angry & bride \\
	\bottomrule
	\end{tabular}
    	\caption{Top 5 Nearest Neighbors of Words followed by Neighbors 45-50 for different Models.}
    \label{nearest_neighbors}
\end{table*}%
\subsection{Training Word Embeddings from scratch}
In this experiment, we empirically observe that training word embeddings using ACE converges significantly faster than NCE after one epoch. As shown in Fig. 3 both ACE (a mixture of $p_{nce}$ and $g_\theta$) and just $g_\theta$ (denoted by ADV) significantly outperforms the NCE baseline, with an absolute improvement of $73.1\%$ and $58.5\%$ respectively on RW score. We note similar results on WordSim-353 dataset where ACE and ADV outperforms NCE by $40.4\%$ and $45.7\%$. We also evaluate our model qualitatively by inspecting the nearest neighbors of selected words in Table.\ \ref{nearest_neighbors}. We first present the five nearest neighbors to each word to show that both NCE and ACE models learn sensible embeddings. We then show that ACE embeddings have much better semantic relevance in a larger neighborhood (nearest neighbor 45-50).
\subsection{Finetuning Word Embeddings}
We take off-the-shelf pre-trained Glove embeddings which were trained using 6 billion tokens \cite{pennington2014glove} and fine-tune them using our algorithm. It is interesting to note that the original Glove objective does not fit into the contrastive learning framework, but nonetheless we find that they benefit from ACE. In fact, we observe that training such that $75\%$ of the words appear as positive contexts is sufficient to beat the largest dimensionality pre-trained Glove model on word similarity tasks. We evaluate our performance on the Rare Word and WordSim353 data. As can be seen from our results in Table \ref{rare_word_tb}, ACE on RW is not always better and for the 100d and 300d Glove embeddings is marginally worse. However, on WordSim353 ACE does considerably better across the board to the point where 50d Glove embeddings outperform the 300d baseline Glove model.

\begin{table}[h]

	\centering
    \resizebox{\columnwidth}{!}{%
	\begin{tabular}[t]{lllllll}
\hline
	& \quad RW & WS353\\
\hline
	Skipgram Only NCE baseline & \quad 18.90 & 31.35\\
    \midrule
    Skipgram + Only ADV  & \quad 29.96 & 58.05 \\
    \midrule
	Skipgram + ACE & \quad 32.71 & 55.00\\
    \midrule
	Glove-50 (Recomputed based on\cite{pennington2014glove}) & \quad 34.02 & 49.51\\
    \midrule
	Glove-100 (Recomputed based on\cite{pennington2014glove})  & \quad 36.64 & 52.76 \\
    \midrule
    Glove-300 (Recomputed based on\cite{pennington2014glove})  & \quad \textbf{41.18} & 60.12\\
    \midrule
    Glove-50 + ACE & \quad 35.60 & 60.46 \\
    \midrule
	Glove-100 + ACE  & \quad 36.51 & 63.29\\
    \midrule
    Glove-300 + ACE  & \quad  40.57 & \textbf{66.50} \\
	\bottomrule
	\end{tabular}%
}
	\caption{Spearman score ($\rho*100$) on RW and WS353 Datasets. We trained a skipgram model from scratch under various settings for only $1$ epoch on wikipedia. For finetuned models we recomputed the scores based on the publicly available 6B tokens Glove models and we finetuned until roughly $75\%$ of the vocabulary was seen. \label{rare_word_tb}}

\end{table}%
\subsection{Hypernym Prediction}
As shown in Table \ref{tbl:Order_Performance}, with ACE training, our method achieves a $1.5\%$ improvement on accuracy over \citet{OrderEmbedding15} without tunning any of the discriminator's hyperparameters. We further report training curve in Fig. 1, we report loss curve on randomly sampled pairs. We stress that in the ACE model, we train random pairs and generator generated pairs jointly, as shown in Fig. 2, hard negatives help the order embedding model converges faster.\\
\begin{table}[h!]
\begin{center}
{
\addtolength{\tabcolsep}{2.5pt}

\resizebox{\columnwidth}{!}{%
\begin{tabular}{|l|c|}
\hline
Method & Accuracy (\%) \\
\hline 
\hline
order-embeddings  &  90.6 \\
order-embeddings + Our ACE  &  \bf{92.0} \\
\hline

\end{tabular}%
}
\caption{Order Embedding Performance}
\label{tbl:Order_Performance}
}
\end{center}
\end{table}
\subsection{Ablation Study and Improving TransD}
\label{sec:wn}
\begin{figure}[h]
\includegraphics[width=\linewidth]{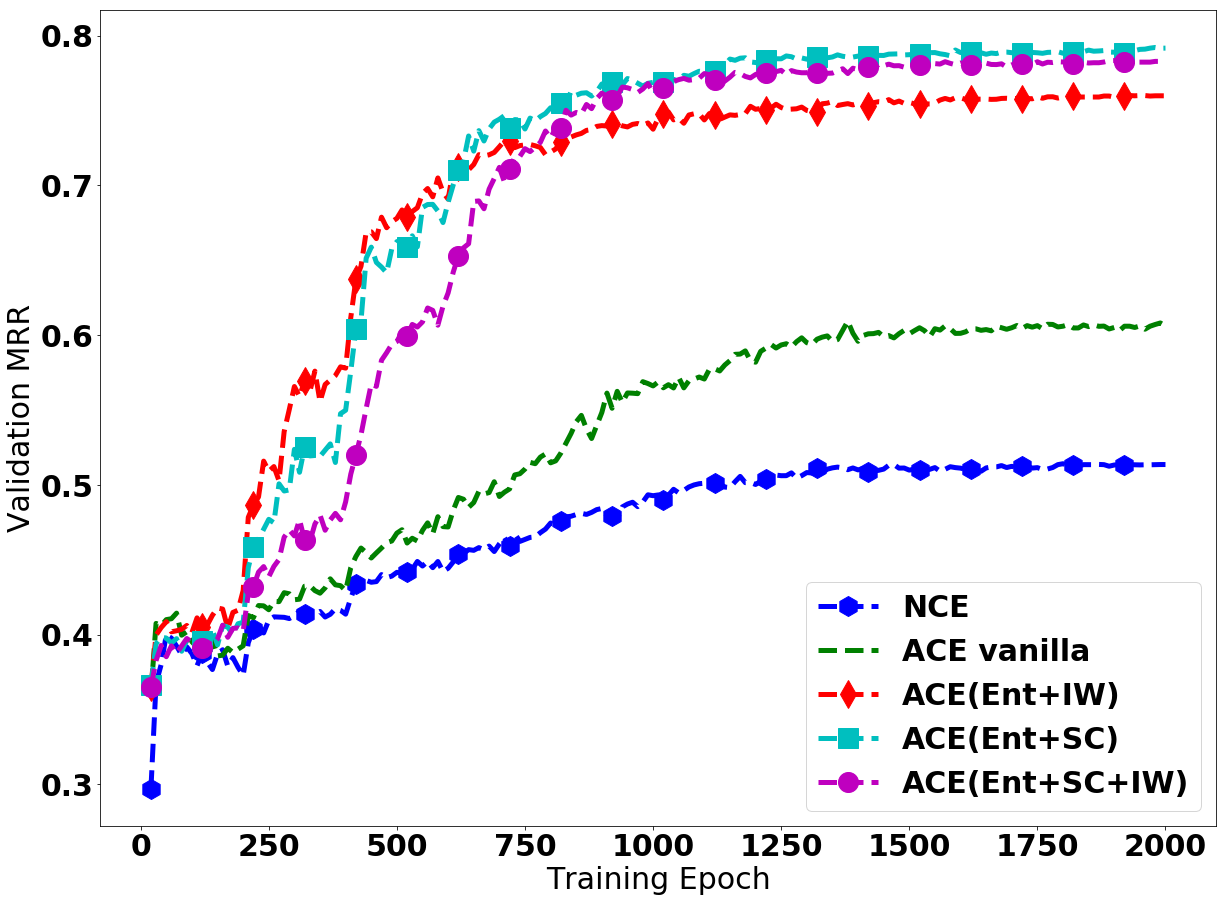}
\caption{Ablation study: measuring validation Mean Reciprocal Rank (MRR) on WN18 dataset as training progresses.}
\label{fig:ablation}
\end{figure}

To analyze different aspects of ACE, we perform an ablation study on the knowledge graph embedding task. As described in Sec.\ \ref{sec:kg}, the base model (discriminator) we apply ACE to is TransD \cite{transD}. Fig.\ \ref{fig:ablation} shows validation performance as training progresses. All variants of ACE converges to better results than base NCE. Among ACE variants, all methods that include entropy regularization significantly outperform without entropy regularization. Without the self critical baseline variance reduction, learning could progress faster at the beginning but the final performance suffers slightly. The best performance is obtained without the additional off-policy learning of the generator. 

Table.\ \ref{table:WN} shows the final test results on WN18 link prediction task.  It is interesting to note that ACE improves MRR score more significantly than hit@10. As MRR is a lot more sensitive to the top rankings, i.e., how the correct configuration ranks among the competitive alternatives, this is consistent with the fact that ACE samples hard negatives and forces the base model to learn a more discriminative representation of the positive examples. 

\begin{table}[h]
\resizebox{\columnwidth}{!}{
\begin{tabular}{lrr}
\toprule
{} &    MRR &  hit@10 \\
\midrule
ACE(Ent+SC)                                  &  \underline{0.792} &   0.945 \\
ACE(Ent+SC+IW)                               &  0.768 &   \underline{\bf{0.949}} \\
NCE TransD (ours)                            &  0.527 &   0.947 \\
NCE TransD (\cite{transD})                   &    - &   0.925 \\
KBGAN(DISTMULT) (\cite{cai2017kbgan}) &  0.772 &   0.948 \\
KBGAN(COMPLEX) (\cite{cai2017kbgan})  &  0.779 &   0.948 \\
Wang et al. (\cite{wang2018incorporating}) & - & 0.93 \\
\midrule
COMPLEX (\cite{complex})  &  \bf{0.941} &   0.947 \\
\bottomrule
\end{tabular}}
\caption{WN18 experiments: the first portion of the table contains results where the base model is TransD, the last separated line is the COMPLEX embedding model \cite{complex}, which achieves the SOTA on this dataset. Among all TransD based models (the best results in this group is underlined), ACE improves over basic NCE and another GAN based approach KBGAN. The gap on MRR is likely due to the difference between TransD and COMPLEX models.}
\vspace{-5mm}
\label{table:WN}
\end{table}

\subsection{Hard Negative Analysis}
To better understand the effect of the adversarial samples proposed by the generator we plot the discriminator loss on both $p_{nce}$ and $g_\theta$ samples. In this context, a harder sample means a higher loss assigned by the discriminator. Fig. 4 shows that discriminator loss for the word embedding task on $g_\theta$ samples are always higher than on $p_{nce}$ samples, confirming that the generator is indeed sampling harder negatives. \\
For Hypernym Prediction task, Fig.\ref{fig:negative_pairs_loss_order_plot} shows discriminator loss on negative pairs sampled from NCE and ACE respectively. The higher the loss the harder the negative pair is. As indicated in the left plot, loss on the ACE negative terms collapses faster than on the NCE negatives. After adding entropy regularization and weight decay, the generator works as expected.
\section{Limitations}
When the generator softmax is large, the current implementation of ACE training is computationally expensive. Although ACE converges faster per iteration, it may converge more slowly on wall-clock time depending on the cost of the softmax. However, embeddings are typically used as pre-trained building blocks for subsequent tasks. Thus, their learning is usually the pre-computation step for the more complex downstream models and spending more time is justified, especially with GPU acceleration. We believe that the computational cost could potentially be reduced via some existing techniques such as the ``augment and reduce" variational inference of \cite{ruiz2018augment}, adaptive softmax \cite{grave2016efficient}, or the ``sparsely-gated" softmax of \citet{shazeer2017outrageously}, but leave that to future work. 

Another limitation is on the theoretical front. As noted in \citet{goodfellow2014distinguishability}, GAN learning does not implement maximum likelihood estimation (MLE), while NCE has MLE as an asymptotic limit. To the best of our knowledge, more distant connections between GAN and MLE training are not known, and tools for analyzing the equilibrium of a min-max game where players are parametrized by deep neural nets are currently not available to the best of our knowledge. 


\section{Conclusion}
In this paper, we propose Adversarial Contrastive Estimation as a general technique for improving supervised learning problems that learn by contrasting observed and fictitious samples. Specifically, we use a generator network in a conditional GAN like setting to propose hard negative examples for our discriminator model. We find that a mixture distribution of randomly sampling negative examples along with an adaptive negative sampler leads to improved performances on a variety of embedding tasks. We validate our hypothesis that hard negative examples are critical to optimal learning and can be proposed via our ACE framework. Finally, we find that controlling the entropy of the generator through a regularization term and properly handling false negatives is crucial for successful training.

\section*{Acknowledgments}
We would like to thank Teng Long for providing the initial baseline code on knowledge graph embeddings, Matthew E. Taylor for proofreading the manuscript and Jackie Chi Kit Cheung for suggestions on preparing for the ACL oral presentation. Additionally, we would also like to acknowledge Jordana Feldman for editing a blog post on this work and April Cooper for creating the artworks for the blog post. Finally, we appreciate the broader Borealis AI team for discussion and emotional support. 

\bibliography{acl2018}

\begin{thebibliography}{50}
\expandafter\ifx\csname natexlab\endcsname\relax\def\natexlab#1{#1}\fi

\bibitem[{Arjovsky et~al.(2017)Arjovsky, Chintala, and
  Bottou}]{arjovsky2017wasserstein}
Martin Arjovsky, Soumith Chintala, and L{\'e}on Bottou. 2017.
\newblock Wasserstein {GAN}.
\newblock \emph{arXiv preprint arXiv:1701.07875}.

\bibitem[{Belanger and McCallum(2016)}]{belanger2016structured}
David Belanger and Andrew McCallum. 2016.
\newblock Structured prediction energy networks.
\newblock In \emph{International Conference on Machine Learning}, pages
  983--992.

\bibitem[{Bordes et~al.(2014)Bordes, Glorot, Weston, and
  Bengio}]{bordes2014semantic}
Antoine Bordes, Xavier Glorot, Jason Weston, and Yoshua Bengio. 2014.
\newblock A semantic matching energy function for learning with
  multi-relational data.
\newblock \emph{Machine Learning}, 94(2):233--259.

\bibitem[{Bordes et~al.(2013)Bordes, Usunier, Garcia-Duran, Weston, and
  Yakhnenko}]{transe}
Antoine Bordes, Nicolas Usunier, Alberto Garcia-Duran, Jason Weston, and Oksana
  Yakhnenko. 2013.
\newblock Translating embeddings for modeling multi-relational data.
\newblock In \emph{Advances in neural information processing systems}, pages
  2787--2795.

\bibitem[{Cai and Wang(2017)}]{cai2017kbgan}
Liwei Cai and William~Yang Wang. 2017.
\newblock Kbgan: Adversarial learning for knowledge graph embeddings.
\newblock \emph{arXiv preprint arXiv:1711.04071}.

\bibitem[{Cao et~al.(2018)Cao, Ding, Lui, and Huang}]{Cao2018Improving}
Yanshuai Cao, Gavin~Weiguang Ding, Kry Yik-Chau Lui, and Ruitong Huang. 2018.
\newblock \href {https://openreview.net/forum?id=BkLhaGZRW} {Improving {GAN}
  training via binarized representation entropy ({BRE}) regularization}.
\newblock In \emph{International Conference on Learning Representations}.

\bibitem[{Dai and Lin(2017)}]{dai2017contrastive}
Bo~Dai and Dahua Lin. 2017.
\newblock Contrastive learning for image captioning.
\newblock In \emph{Advances in Neural Information Processing Systems}, pages
  898--907.

\bibitem[{Dettmers et~al.(2017)Dettmers, Minervini, Stenetorp, and
  Riedel}]{convE}
Tim Dettmers, Pasquale Minervini, Pontus Stenetorp, and Sebastian Riedel. 2017.
\newblock Convolutional 2d knowledge graph embeddings.
\newblock \emph{arXiv preprint arXiv:1707.01476}.

\bibitem[{Dyer(2014)}]{dyer2014notes}
Chris Dyer. 2014.
\newblock Notes on noise contrastive estimation and negative sampling.
\newblock \emph{arXiv preprint arXiv:1410.8251}.

\bibitem[{Fedus et~al.(2018)Fedus, Goodfellow, and Dai}]{fedus2018maskgan}
William Fedus, Ian Goodfellow, and Andrew~M Dai. 2018.
\newblock Mask{GAN}: Better text generation via filling in the \_.
\newblock \emph{arXiv preprint arXiv:1801.07736}.

\bibitem[{Finkelstein et~al.(2001)Finkelstein, Gabrilovich, Matias, Rivlin,
  Solan, Wolfman, and Ruppin}]{finkelstein2001placing}
Lev Finkelstein, Evgeniy Gabrilovich, Yossi Matias, Ehud Rivlin, Zach Solan,
  Gadi Wolfman, and Eytan Ruppin. 2001.
\newblock Placing search in context: The concept revisited.
\newblock In \emph{Proceedings of the 10th international conference on World
  Wide Web}, pages 406--414. ACM.

\bibitem[{Goodfellow et~al.(2014{\natexlab{a}})Goodfellow, Pouget-Abadie,
  Mirza, Xu, Warde-Farley, Ozair, Courville, and Bengio}]{NIPS2014_5423}
Ian Goodfellow, Jean Pouget-Abadie, Mehdi Mirza, Bing Xu, David Warde-Farley,
  Sherjil Ozair, Aaron Courville, and Yoshua Bengio. 2014{\natexlab{a}}.
\newblock \href
  {http://papers.nips.cc/paper/5423-generative-adversarial-nets.pdf}
  {Generative adversarial nets}.
\newblock In Z.~Ghahramani, M.~Welling, C.~Cortes, N.~D. Lawrence, and K.~Q.
  Weinberger, editors, \emph{Advances in Neural Information Processing Systems
  27}, pages 2672--2680.

\bibitem[{Goodfellow et~al.(2014{\natexlab{b}})Goodfellow, Pouget-Abadie,
  Mirza, Xu, Warde-Farley, Ozair, Courville, and
  Bengio}]{goodfellow2014generative}
Ian Goodfellow, Jean Pouget-Abadie, Mehdi Mirza, Bing Xu, David Warde-Farley,
  Sherjil Ozair, Aaron Courville, and Yoshua Bengio. 2014{\natexlab{b}}.
\newblock Generative adversarial nets.
\newblock In \emph{Advances in neural information processing systems}, pages
  2672--2680.

\bibitem[{Goodfellow(2014)}]{goodfellow2014distinguishability}
Ian~J Goodfellow. 2014.
\newblock On distinguishability criteria for estimating generative models.
\newblock \emph{arXiv preprint arXiv:1412.6515}.

\bibitem[{Grathwohl et~al.(2017)Grathwohl, Choi, Wu, Roeder, and
  Duvenaud}]{grathwohl2017backpropagation}
Will Grathwohl, Dami Choi, Yuhuai Wu, Geoff Roeder, and David Duvenaud. 2017.
\newblock Backpropagation through the void: Optimizing control variates for
  black-box gradient estimation.
\newblock \emph{arXiv preprint arXiv:1711.00123}.

\bibitem[{Grave et~al.(2016)Grave, Joulin, Ciss{\'e}, Grangier, and
  J{\'e}gou}]{grave2016efficient}
Edouard Grave, Armand Joulin, Moustapha Ciss{\'e}, David Grangier, and
  Herv{\'e} J{\'e}gou. 2016.
\newblock Efficient softmax approximation for {GPU}s.
\newblock \emph{arXiv preprint arXiv:1609.04309}.

\bibitem[{Gulrajani et~al.(2017)Gulrajani, Ahmed, Arjovsky, Dumoulin, and
  Courville}]{gulrajani2017improved}
Ishaan Gulrajani, Faruk Ahmed, Martin Arjovsky, Vincent Dumoulin, and Aaron~C
  Courville. 2017.
\newblock Improved training of wasserstein gans.
\newblock In \emph{Advances in Neural Information Processing Systems}, pages
  5769--5779.

\bibitem[{Gutmann and Hyv{\"a}rinen(2010)}]{gutmann2010noise}
Michael Gutmann and Aapo Hyv{\"a}rinen. 2010.
\newblock Noise-contrastive estimation: A new estimation principle for
  unnormalized statistical models.
\newblock In \emph{Proceedings of the Thirteenth International Conference on
  Artificial Intelligence and Statistics}, pages 297--304.

\bibitem[{Gutmann and Hyv{\"a}rinen(2012)}]{gutmann2012noise}
Michael~U Gutmann and Aapo Hyv{\"a}rinen. 2012.
\newblock Noise-contrastive estimation of unnormalized statistical models, with
  applications to natural image statistics.
\newblock \emph{Journal of Machine Learning Research}, 13(Feb):307--361.

\bibitem[{Jang et~al.(2016)Jang, Gu, and Poole}]{jang2016categorical}
Eric Jang, Shixiang Gu, and Ben Poole. 2016.
\newblock Categorical reparameterization with gumbel-softmax.
\newblock \emph{arXiv preprint arXiv:1611.01144}.

\bibitem[{Ji et~al.(2015)Ji, He, Xu, Liu, and Zhao}]{transD}
Guoliang Ji, Shizhu He, Liheng Xu, Kang Liu, and Jun Zhao. 2015.
\newblock Knowledge graph embedding via dynamic mapping matrix.
\newblock In \emph{Proceedings of the 53rd Annual Meeting of the Association
  for Computational Linguistics and the 7th International Joint Conference on
  Natural Language Processing (Volume 1: Long Papers)}, volume~1, pages
  687--696.

\bibitem[{Kingma and Ba(2014)}]{kingma2014adam}
Diederik~P Kingma and Jimmy Ba. 2014.
\newblock Adam: A method for stochastic optimization.
\newblock \emph{arXiv preprint arXiv:1412.6980}.

\bibitem[{Lin et~al.(2015)Lin, Liu, Sun, Liu, and Zhu}]{lin2015learning}
Yankai Lin, Zhiyuan Liu, Maosong Sun, Yang Liu, and Xuan Zhu. 2015.
\newblock Learning entity and relation embeddings for knowledge graph
  completion.
\newblock In \emph{AAAI}, volume~15, pages 2181--2187.

\bibitem[{Liu et~al.(2018)Liu, Feng, Mao, Zhou, Peng, and
  Liu}]{liu2018actiondependent}
Hao Liu, Yihao Feng, Yi~Mao, Dengyong Zhou, Jian Peng, and Qiang Liu. 2018.
\newblock \href {https://openreview.net/forum?id=H1mCp-ZRZ} {Action-dependent
  control variates for policy optimization via stein identity}.
\newblock In \emph{International Conference on Learning Representations}.

\bibitem[{Luong et~al.(2013)Luong, Socher, and Manning}]{luong2013better}
Thang Luong, Richard Socher, and Christopher~D Manning. 2013.
\newblock Better word representations with recursive neural networks for
  morphology.
\newblock In \emph{CoNLL}, pages 104--113.

\bibitem[{Maddison et~al.(2016)Maddison, Mnih, and Teh}]{maddison2016concrete}
Chris~J Maddison, Andriy Mnih, and Yee~Whye Teh. 2016.
\newblock The concrete distribution: A continuous relaxation of discrete random
  variables.
\newblock \emph{arXiv preprint arXiv:1611.00712}.

\bibitem[{Mikolov et~al.(2013)Mikolov, Sutskever, Chen, Corrado, and
  Dean}]{mikolov2013distributed}
Tomas Mikolov, Ilya Sutskever, Kai Chen, Greg~S Corrado, and Jeff Dean. 2013.
\newblock Distributed representations of words and phrases and their
  compositionality.
\newblock In \emph{Advances in neural information processing systems}, pages
  3111--3119.

\bibitem[{{Mirza} and {Osindero}(2014)}]{ConditionalGan}
M.~{Mirza} and S.~{Osindero}. 2014.
\newblock \href {http://arxiv.org/abs/1411.1784} {{Conditional Generative
  Adversarial Nets}}.
\newblock \emph{ArXiv e-prints}.

\bibitem[{Mnih and Kavukcuoglu(2013)}]{mnih2013learning}
Andriy Mnih and Koray Kavukcuoglu. 2013.
\newblock Learning word embeddings efficiently with noise-contrastive
  estimation.
\newblock In \emph{Advances in neural information processing systems}, pages
  2265--2273.

\bibitem[{Mnih and Teh(2012)}]{mnih2012fast}
Andriy Mnih and Yee~Whye Teh. 2012.
\newblock A fast and simple algorithm for training neural probabilistic
  language models.
\newblock \emph{arXiv preprint arXiv:1206.6426}.

\bibitem[{Pennington et~al.(2014)Pennington, Socher, and
  Manning}]{pennington2014glove}
Jeffrey Pennington, Richard Socher, and Christopher Manning. 2014.
\newblock Glove: Global vectors for word representation.
\newblock In \emph{Proceedings of the 2014 conference on empirical methods in
  natural language processing (EMNLP)}, pages 1532--1543.

\bibitem[{Rennie et~al.(2016)Rennie, Marcheret, Mroueh, Ross, and
  Goel}]{rennie2016self}
Steven~J Rennie, Etienne Marcheret, Youssef Mroueh, Jarret Ross, and Vaibhava
  Goel. 2016.
\newblock Self-critical sequence training for image captioning.
\newblock \emph{arXiv preprint arXiv:1612.00563}.

\bibitem[{Ruiz et~al.(2018)Ruiz, Titsias, Dieng, and Blei}]{ruiz2018augment}
Francisco~JR Ruiz, Michalis~K Titsias, Adji~B Dieng, and David~M Blei. 2018.
\newblock Augment and reduce: Stochastic inference for large categorical
  distributions.
\newblock \emph{arXiv preprint arXiv:1802.04220}.

\bibitem[{Schroff et~al.(2015)Schroff, Kalenichenko, and
  Philbin}]{schroff2015facenet}
Florian Schroff, Dmitry Kalenichenko, and James Philbin. 2015.
\newblock Facenet: A unified embedding for face recognition and clustering.
\newblock In \emph{Proceedings of the IEEE Conference on Computer Vision and
  Pattern Recognition}, pages 815--823.

\bibitem[{Shazeer et~al.(2017)Shazeer, Mirhoseini, Maziarz, Davis, Le, Hinton,
  and Dean}]{shazeer2017outrageously}
Noam Shazeer, Azalia Mirhoseini, Krzysztof Maziarz, Andy Davis, Quoc Le,
  Geoffrey Hinton, and Jeff Dean. 2017.
\newblock Outrageously large neural networks: The sparsely-gated
  mixture-of-experts layer.
\newblock \emph{arXiv preprint arXiv:1701.06538}.

\bibitem[{Shrivastava et~al.(2016)Shrivastava, Gupta, and
  Girshick}]{shrivastava2016training}
Abhinav Shrivastava, Abhinav Gupta, and Ross Girshick. 2016.
\newblock Training region-based object detectors with online hard example
  mining.
\newblock In \emph{Proceedings of the IEEE Conference on Computer Vision and
  Pattern Recognition}, pages 761--769.

\bibitem[{Smith and Eisner(2005)}]{smith2005contrastive}
Noah~A Smith and Jason Eisner. 2005.
\newblock Contrastive estimation: Training log-linear models on unlabeled data.
\newblock In \emph{Proceedings of the 43rd Annual Meeting on Association for
  Computational Linguistics}, pages 354--362. Association for Computational
  Linguistics.

\bibitem[{Taskar et~al.(2005)Taskar, Chatalbashev, Koller, and
  Guestrin}]{taskar2005learning}
Ben Taskar, Vassil Chatalbashev, Daphne Koller, and Carlos Guestrin. 2005.
\newblock Learning structured prediction models: A large margin approach.
\newblock In \emph{Proceedings of the 22nd international conference on Machine
  learning}, pages 896--903. ACM.

\bibitem[{Trouillon et~al.(2016)Trouillon, Welbl, Riedel, Gaussier, and
  Bouchard}]{complex}
Th{\'e}o Trouillon, Johannes Welbl, Sebastian Riedel, {\'E}ric Gaussier, and
  Guillaume Bouchard. 2016.
\newblock Complex embeddings for simple link prediction.
\newblock In \emph{International Conference on Machine Learning}, pages
  2071--2080.

\bibitem[{Tsochantaridis et~al.(2005)Tsochantaridis, Joachims, Hofmann, and
  Altun}]{tsochantaridis2005large}
Ioannis Tsochantaridis, Thorsten Joachims, Thomas Hofmann, and Yasemin Altun.
  2005.
\newblock Large margin methods for structured and interdependent output
  variables.
\newblock \emph{Journal of machine learning research}, 6(Sep):1453--1484.

\bibitem[{Tu and Gimpel(2018)}]{tu2018learning}
Lifu Tu and Kevin Gimpel. 2018.
\newblock Learning approximate inference networks for structured prediction.
\newblock In \emph{International Conference on Learning Representations}.

\bibitem[{Tucker et~al.(2017)Tucker, Mnih, Maddison, Lawson, and
  Sohl-Dickstein}]{NIPS2017_6856}
George Tucker, Andriy Mnih, Chris~J Maddison, John Lawson, and Jascha
  Sohl-Dickstein. 2017.
\newblock \href
  {http://papers.nips.cc/paper/6856-rebar-low-variance-unbiased-gradient-estimates-for-discrete-latent-variable-models.pdf}
  {Rebar: Low-variance, unbiased gradient estimates for discrete latent
  variable models}.
\newblock In I.~Guyon, U.~V. Luxburg, S.~Bengio, H.~Wallach, R.~Fergus,
  S.~Vishwanathan, and R.~Garnett, editors, \emph{Advances in Neural
  Information Processing Systems 30}, pages 2627--2636. Curran Associates, Inc.

\bibitem[{Vaswani et~al.(2013)Vaswani, Zhao, Fossum, and
  Chiang}]{vaswani2013decoding}
Ashish Vaswani, Yinggong Zhao, Victoria Fossum, and David Chiang. 2013.
\newblock Decoding with large-scale neural language models improves
  translation.
\newblock In \emph{Proceedings of the 2013 Conference on Empirical Methods in
  Natural Language Processing}, pages 1387--1392.

\bibitem[{Vendrov et~al.(2016)Vendrov, Kiros, Fidler, and
  Urtasun}]{OrderEmbedding15}
Ivan Vendrov, Ryan Kiros, Sanja Fidler, and Raquel Urtasun. 2016.
\newblock Order-embeddings of images and language.
\newblock In \emph{International Conference on Learning Representations}.

\bibitem[{Wang et~al.(2018)Wang, Li, and Pan}]{wang2018incorporating}
Peifeng Wang, Shuangyin Li, and Rong Pan. 2018.
\newblock Incorporating {GAN} for negative sampling in knowledge representation
  learning.
\newblock In \emph{The Thirty-Second AAAI Conference on Artificial Intelligence
  (AAAI-18)}.

\bibitem[{Wang et~al.(2014)Wang, Zhang, Feng, and Chen}]{transH}
Zhen Wang, Jianwen Zhang, Jianlin Feng, and Zheng Chen. 2014.
\newblock Knowledge graph embedding by translating on hyperplanes.
\newblock In \emph{Proceedings of the Twenty-Eighth AAAI Conference on
  Artificial Intelligence}, pages 1112--1119. AAAI Press.

\bibitem[{Williams(1992)}]{williams1992simple}
Ronald~J Williams. 1992.
\newblock Simple statistical gradient-following algorithms for connectionist
  reinforcement learning.
\newblock \emph{Machine learning}, 8(3-4):229--256.

\bibitem[{Yang et~al.(2014)Yang, Yih, He, Gao, and Deng}]{distmult}
Bishan Yang, Wen-tau Yih, Xiaodong He, Jianfeng Gao, and Li~Deng. 2014.
\newblock Embedding entities and relations for learning and inference in
  knowledge bases.
\newblock \emph{arXiv preprint arXiv:1412.6575}.

\bibitem[{Yoshua et~al.(2003)Yoshua, Rejean, Pascal, and
  ´~Christian}]{Yoshua2003}
Bengio Yoshua, Ducharme Rejean, Vincent Pascal, and Jauvin ´~Christian. 2003.
\newblock A neural probabilistic language model.
\newblock \emph{Journal of Machine Learning Research.}

\bibitem[{Zhao et~al.(2016)Zhao, Mathieu, and LeCun}]{zhao2016energy}
Junbo Zhao, Michael Mathieu, and Yann LeCun. 2016.
\newblock Energy-based generative adversarial network.
\newblock \emph{arXiv preprint arXiv:1609.03126}.

\end{thebibliography}
\bibliographystyle{acl_natbib}
\end{document}